\documentclass[sigconf,screen]{acmart}

\AtBeginDocument{%
  }

\setcopyright{rightsretained}
\copyrightyear{2024}
\acmYear{2024}

\usepackage{hyperref}
\usepackage{hyperxmp}
\usepackage{graphicx}
\definecolor{addedtext}{rgb}{1, 0, 0}

\acmConference[]{}{2024}{}

\begin{document}

\title{MiningGPT - A Domain-Specific Large Language Model \\ for the Mining Industry}

\author{Kurukulasooriya Fernando}
\affiliation{%
  \institution{University of Queensland}
  \city{St. Lucia, QLD}
  \country{Australia}}
\email{kurukulasooriya.fernando@student.uq.edu.au}

\author{Gianluca Demartini}
\affiliation{%
  \institution{University of Queensland}
  \city{St. Lucia, QLD}
  \country{Australia}}
\email{g.demartini@uq.edu.au}


\begin{abstract}
 Recent advancements of generative LLMs (Large Language Models) have exhibited human-like language capabilities but have shown a lack of domain-specific understanding. Therefore, the research community has started the development of domain-specific LLMs for many domains. In this work we focus on discussing how to build mining domain-specific LLMs, as the global mining industry contributes significantly to the worldwide economy. 
 
We report on MiningGPT, a mining domain-specific instruction-following 7B parameter LLM model which showed a 14\% higher mining domain knowledge test score as compared to its parent model Mistral 7B instruct. 

\end{abstract}

\begin{teaserfigure}
\centering
  \includegraphics[width=0.8\textwidth]{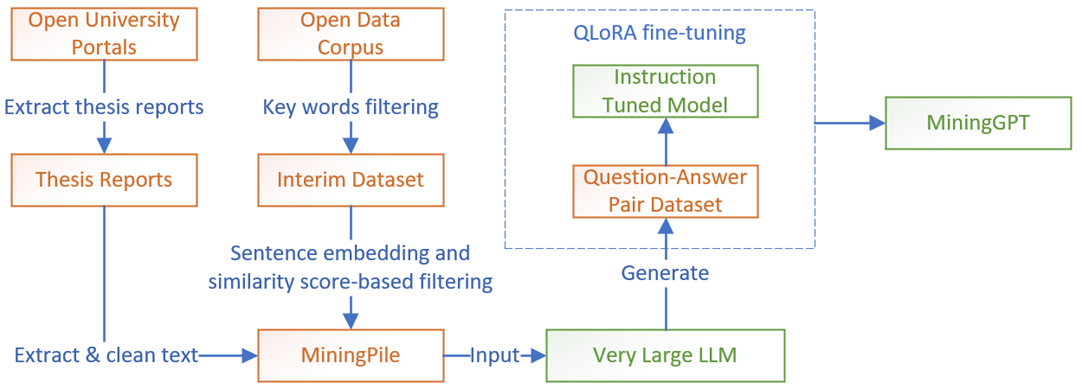}
  \caption{The approach used to build MiningGPT.}
  \Description{The approach used to build MiningGPT.}
  \label{fig:teaser}
\end{teaserfigure}


\maketitle

\section{Introduction}
Recent advancements in instruction-following generative LLMs (Large Language Models) such as ChatGPT have exhibited human-like capabilities in the general domain. However, these models have shown a lack of knowledge in specific domains. Therefore, building such LLMs for specific domains has become necessary to unlock the true economic value of these technological advancements providing a greater benefit to the broader global community.
As the global mining industry contributes significantly to the global economy \cite{HodgeR.Anthony2022Tgmi}, it expected  that building a open-source, domain-specific LLM for the mining industry would have positive economic impacts on the global economy.

In this paper we present the methodology used to build our model as well as experimental evidence aimed at comparatively evaluating its performance as compared to general-domain baselines.

\section{Methodology}


As there was no previous research conducted around building domain-specific LLMs for the mining industry, our open research question in this work is:
\begin{itemize}
    \item Would a mining industry domain-specific LLM perform better than a foundational LLM on common-sense reasoning and question answering tasks?
\end{itemize}

Based on the insights from our  review of the literature, the following research methodology was iteratively developed and followed to investigate the research question in a step-by-step manner.

Study 1 - Creation of a mining domain-specific open dataset
\begin{itemize}
\item Compiled a comprehensive mining domain-specific terminology dataset and a domain-specific body of knowledge reference datasets.
\item Mined open datasets, which include mainly the C4 corpus \cite{DodgeJesse2021DLWC} and arXiv papers using keyword extraction and a combination of sentence embedding methods and a similarity score-based approach to build a mining domain-specific open dataset. The dataset was named MiningPile.
\item Handpicked mining domain-specific thesis reports from open university portals, transformed them into a tabular format and added them to the MiningPile.
\end{itemize}

Study 2 - Building a mining domain-specific language model
\begin{itemize}
\item Built a question-answer dataset from MiningPile by employing a very large (> 100B parameters) general-domain LLM. 
\item Fine-tuned an instruction-tuned foundational model using the question-answer dataset. The fine-tuned model was named MiningGPT.
\item Evaluated and compared the performance of the fine-tuned model against other open-source models.
\end{itemize}

Google Colaboratory, also known as Colab, was used as the model-building platform and PyTorch was used as the development framework. The libraries provided under the Hugging Face ecosystem and the Unsloth library were utilised for the model training.

In the rest of the paper we present the the dataset construction approach with statistics of the final datasets together with an experimental evaluation of MiningGPT as compared to a general-domain model.

\section{Study 1 – Building MiningPile}
The first study conducts a series of experiments and builds a comprehensive mining industry domain-specific open dataset that can be used for building a mining domain-specific large language model.
\subsection{Methodology}
\subsubsection{Keyword-Based Filtering of Open Datasets}
Based on existing literature on domain-specific LLMs, the primary method used for  building  MiningPile was to filter open large datasets using keywords.
The mining industry is very broad and spans operations and maintenance, finance, marketing, supply chain, human resources, law, etc. However, as operation and maintenance are the backbone of the industry, the data filtering was confined to this type of data. 
As the first step, a comprehensive glossary with domain-specific terminology was created using domain expertise. The glossary consists of 600 keywords related to the mining domain and careful consideration was given to ensure the keywords were contemporary.
The keywords were used for the filtering of open datasets (available on the Hugging Face platform). Datasets were processed in batches of 1 million samples.

\subsubsection{Further Filtering of Datasets Using a Combination of Sentence Embedding and Similarity Score-based Methods}
The filtered dataset was examined for content quality of, based on domain expert judgment, and it was revealed that the dataset contained data from domains other than mining. Further investigation revealed that some of those keywords are used in other domains as well.
Therefore, a combination of sentence embedding, and cosine similarity score-based method was developed and used for the further filtering of the datasets. The steps of the method are as follows.

\begin{itemize}
\item Build embeddings for a reference knowledge dataset.
\item Calculate the embedding vectors for the text field of each row of the dataset.
\item For each row, find the embedding vectors most similar to the text field from the list of embeddings of the reference knowledge dataset and its score.
\item For each row, add the max similarity score and the index of the relevant row of the reference knowledge dataset into new columns.
\end{itemize}

As the first step of the implementation of the above method, a domain-specific body of knowledge (i.e., the reference knowledge dataset) was compiled referring to literature on the open web. The pre-trained transformer model ``sentence-transformers/multi-qa-mpnet-base-dot-v1" (used for calculating word embeddings) provided by the Transformers library of Hugging Face and the FAISS library\footnote{\url{https://github.com/facebookresearch/faiss}}  was used for the implementation of the method.

\subsubsection{Improving Processing Efficiency of Further Filtering Algorithm}
\begin{figure}[h]
  \centering
  \includegraphics[width=\linewidth]{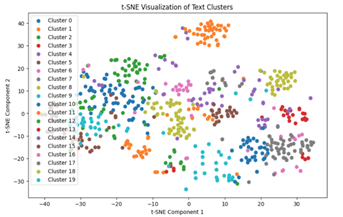}
  \caption{Clustering of the sentence embedding of the reference knowledge dataset}
  \Description{clustering of the sentence embedding of the reference knowledge dataset}
  \label{fig:clustering}
\end{figure}

We used data clustering methods to reduce the size of the reference knowledge dataset thus improving data processing efficiency for the above method. The visualisation of the clustering of the sentence embedding of the reference knowledge dataset, using  k-means and visualised in a 2-dimensional space using principal component analysis \cite{GewersFelipeL.2021PcaA} is shown in Figure \ref{fig:clustering}.

The results showed that the size of the dataset could be potentially reduced, as there were clusters of data points, which were very close to each other in the sentence embedding space. Therefore, expert judgement was adopted to find and remove some of the data points based on their similarity score with their nearest neighbour datapoints. This step reduced the size of the reference knowledge dataset by 66.56\%.
Subsequent calculations revealed that the above de-duplication exercise resulted in a 25.21\% reduction in the duration of the further filtering method.

The filtered dataset was further filtered using the above method and resulted in a double-filtered mining domain-specific dataset. The final dataset was selected based on a minimum cut-off value of the max similarity score. Domain expert judgment was used in finding the suitable cut-off mark and selected 0.65 as the suitable cut-off value. Careful consideration was given in selecting the cut-off value, as a very high score could result in a smaller dataset with heavy mining domain-specific content, which would result in model over-fitting issues at the model-building stage.

\subsubsection{Extracting Thesis Reports from Open University Portals}
Thesis reports from open university portals could provide a valuable source for building a mining domain-specific data corpus. Therefore, we built a web crawler (using open-source Python libraries) that could automatically extract thesis reports from open university portals and compile a dataset.
Using that and manually handpicking from 19 university portals, around 1,200 mining domain-specific thesis reports were   added to the MiningPile.

The composition of MiningPile is illustrated in Table \ref{tab:freq}.
\begin{table}
  \caption{Composition of MiningPile.}
  \label{tab:freq}
  \begin{tabular}{ccl}
    \toprule
    Data Source&Count of Rows&Token Count (M)\\
    \midrule
    Open data and books& 76,229&81.8\\
    Thesis reports& 91,628 &38.8\\
    Total& 167,857&120.6\\
\end{tabular}
\end{table}

As a result of an exploratory data analysis, it was found that some of the keywords have a very high representation within the dataset. Table \ref{tab:topn} shows the top 10 keywords and their frequency in the dataset.

\begin{table}
  \caption{Top 10 keywords in MiningPile.}
  \label{tab:topn}
  \begin{tabular}{cl}
    \toprule
    Keyword&Count\\
    \midrule
    Crusher&176,541\\
    Gold&94,313\\
 Coal&69,017\\
 Drill&49,653\\
 Copper&46,107\\
 Conveyor&28,531\\
    slurry&10,206\\
 Leaching&9,189\\
 Classifier&8,875\\
 Blasting&8,120\\
\end{tabular}
\end{table}

\subsection{Discussion}
To assess the properties of  MiningPile, a literature review was done to collect information about the size of the datasets used in building other domain-specific LLMs.
The results show that MiningPile could be used for  model fine-tuning in building a mining domain-specific LLM, as datasets that are 58.3\% smaller \cite{LiYinheng2023LLMi} than MiningPile have been used for model fine-tuning in the finance domain with acceptable levels of performance. 
Therefore, we conclude
that
MiningPile could be used for  model fine-tuning to build domain-specific LLMs in the mining domain.
On the other hand, MiningPile would not be useful for the model pre-training in building a mining domain-specific LLM, as datasets that have been used for the successful model pre-training in the finance domain are significantly larger than MiningPile. 

Also, it was observed that around 8\% of the data rows of C4 contain at least one keyword used in the mining domain. However, it was also observed that only 0.6\% of rows contain mining domain-specific data. This pattern is evenly distributed across the dataset. Intuitively, what LLMs learn during the pre-training process is the associations between words in the training data. Therefore, a foundational model pre-trained on C4 corpus would have already learnt the associations between mining domain-specific keywords.

\section{Study 2 – Building and Evaluating MiningGPT}

\subsection{Methodology}
As per previous study, only model fine-tuning methods were used in building the domain-specific model. As per insights from the literature, decoder-only causal language-based models were considered for the experiment and  LoRA \cite{DettmersTim2023QEFo} was employed for the model fine-tuning as it approximates the full fine-tuning of a foundational model. More specifically, the QLoRA \cite{DettmersTim2023QEFo} method was employed for the experimentation given the widespread availability of existing code bases, and highly optimised supporting libraries.

In addition, various model fine-tuning approaches, used for building domain-specific language models in other domains were investigated. We observed that most of those approaches could be easily applied in building a mining domain-specific LLM as well. However, a novel fine-tuning approach was proposed and experimented to take advantage of the result of the previous study. Based on the previous study, it was hypothesised,

{\bfseries HYPOTHESIS 1:}
An adapter-based fine-tuning of a foundational model, pre-trained on C4 corpus (which has substantially learnt associations between mining domain-related keywords in the embedding space), with a mining domain-specific dataset would enhance the model’s mining domain knowledge, as the fine-tuning process would result in concentrating those associations into the mining domain in embedding space.

In addition, based on the observations of fine-tuning approaches used in other domains and the theory of QLoRA, it was hypothesised,

{\bfseries HYPOTHESIS 2:}
A  QLoRA fine-tuning of an instruction-tuned foundational model with a question-answer pair dataset would not result in catastrophic forgetting of instruction-following capabilities of the model, as new knowledge is being learnt in the adaptor.

Based on the above two hypotheses, an experiment was designed to test a novel fine-tuning approach and the steps of the design of the experiment are as follows.
\begin{itemize}
\item Create a question-answer pair dataset from MiningPile.
\item QLoRA fine-tuning of an instruction-tuned foundational model, which had been pre-trained on a training dataset, which contained the C4 corpus.
\end{itemize}

\subsubsection{Create a Question-Answer Pair Dataset from MiningPile}
A very large language model (i.e., Google Gemini 1.5) was used in generating question-answer pairs using MiningPile as the input. The prompt for the model was carefully crafted in a way that the model generates high-quality question-answer pairs in the operation and maintenance sub-domain in the mining domain. The prompt was presented to the model followed bythe MiningPile dataset.

\subsubsection{QLoRA Fine-tuning of an Instruction-tuned Foundational Model}
As per insights from the literature, the model architecture and training technique were prioritised over model size when adapting LLMs in the mining domain. Therefore, the preference was smaller models with better model architectures. A consideration was also given in choosing the model size, due to  resource constraints. Therefore, a model with a maximum of 7B parameters was identified as the ideal size.
As a result, Mistral 7B Instruct was used for the experimentation as it fulfils all the above requirements \cite{JiangAlbertQ2023M7}. 
%
The hyper-parameters play a critical role in the performance of the model. Therefore, the fine-tuning process was set up as a series of model fine-tuning runs by varying models’ hyper-parameters. Also, the runs were conducted in a way that each run would build upon lessons learnt from the previous runs.

\subsubsection{Mining Domain-Specific Performance Evaluation}
The mining domain-specific performance evaluation was targeted to test the mining domain knowledge of the model. Therefore, the model evaluation dataset, which was created as part of the question-answer pair dataset preparation work, was used for the performance evaluation of the fine-tuned model.
%
As METERO was not suitable as unable to measure the similarity of the `meanings' in the mining domain, we used cosine similarity for the calculation of the similarity of two sentences, as it had already been tested useful during the dataset-building stage. Careful consideration was given in designing the performance evaluation method, in a way that the use of cosine similarity would not dilute the evaluation results.

The evaluation dataset consists of question-and-answer pairs. For each question, the model's output was generated and compared to the corresponding gold answer. During the evaluation process, sentence embeddings were calculated for both the gold answer and the model's output using the pre-trained transformer model ``sentence-transformers/multi-qa-mpnet-base-dot-v1'' from the Transformers library. The cosine similarity between these embeddings was then computed. If the cosine similarity exceeded 0.85, the model's response was considered a correct answer. This high threshold was chosen to ensure that only outputs closely matching the gold answers in the sentence embedding space were counted towards the model's effectiveness, thereby excluding random guesses.

\subsubsection{General Domain Performance Evaluation}
As the model was fine-tuned using an instruction-tuned foundational model, the performance evaluation in the general domain was confined to testing if the fine-tuned model had successfully retained the capabilities of the base model. Therefore, we tested both the fine-tuned model and the base model with datasets such as OpenbookQA, CommonsenseQA, and HellaSwag. For each dataset, we recorded the test scores for both models. Then, for each dataset, we calculated how much the fine-tuned model's score deviated from the base model's score. This deviation was expressed as a percentage of the base model's score.


\subsubsection{Qualitative Performance Evaluation}
A qualitative performance evaluation was conducted to evaluate the quality of the MiningGPT responses. MiningGPT and ChatGPT (get-3.5-turbo) were given a set of questions and the answers were compared for their quality using domain expert judgment. Please see Appendix A for qualitative example results of the model.

\subsection{Results}


\subsubsection{QLoRA Fine-tuning Results}
The fine-tuning of the model was conducted as a series of model fine-tuning runs by varying models’ hyper-parameters. The selected model hyper-parameters are shown in Table
\textcolor{addedtext}{\ref{tab:reshyp}}. The learning rates vs performance of the final evaluated model checkpoints are illustrated in Table \ref{tab:resrate}. 
The checkpoint at the 2nd epoch with a learning rate of 1e-4 was selected as the final model.

\begin{table}
  \caption{The hyperparameters of the selected checkpoint.}
  \label{tab:reshyp}
  \begin{tabular}{ccl}
    \toprule
    Hyperparameter & Value \\
    \midrule
    LoRA rank & 128 \\
    LoRA alpha & 16 \\
    LoRA dropout & 0.01 \\
    Learning rate & 1e-4 \\
    Batch size & 16 \\
    Weight decay & 0.01 \\
    Warmup steps & 5 \\
    Gradient accumulation steps & 4 \\
    \bottomrule
  \end{tabular}
\end{table}

\begin{table}
  \caption{Performance score for different learning rates and model checkpoints.}
  \label{tab:resrate}
  \begin{tabular}{ccccl}
    \toprule
    Check Point & 1e-4 & 2e-4 & 3e-4 \\
    \midrule
    Epoch 1 & 54.8 & 54.23 & 53.5 \\
    Epoch 2 & 55.51 & 55.34 & 55.1 \\
    Epoch 3 & 54.9 & 55.1 & 55.4 \\
    Epoch 4 & 54 & 53.4 & 52 \\
    \bottomrule
  \end{tabular}
\end{table}

\subsubsection{Mining Domain-Specific Performance Evaluation Results}
The mining domain-specific performance evaluation results are illustrated in Table \ref{tab:res}.

\begin{table}
  \caption{Mining domain-specific performance evaluation results.}
  \label{tab:res}
  \begin{tabular}{ccc}
    \toprule
    Model&Size&Performance Score \%\\
    \midrule
    Mininggpt 7b instruct v3& 7B&55.5\\
    Falcon-7b-instruct& 7B&46.5\\
    Mistral-7B-instruct-v0.2& 7B&41.2\\
    Meta-Llama-3-8B-instruct& 8B&35.5\\
\end{tabular}
\end{table}

The performance of the base model that was used for the fine-tuning was evaluated to establish baseline performance. The result showed that MiningGPT is 14 percent ahead of Mistral-7B-Instruct-v0.2 in answering questions in the mining domain. In addition, some other open source LLM were evaluated using the evaluation dataset and compared against the MiningGPT model performance. The result showed that MiningGPT is ahead of all the evaluated models of similar size.

\subsubsection{General Domain Performance Evaluation Results}
The deviation of the MiningGPT score from the Mistral score as a percentage was used as the evaluation metric and the results are shown in Table \ref{tab:genres}. The results were evaluated using the categorisation used in Mistral 7B evaluation \cite{JiangAlbertQ2023M7}.

\begin{table}
  \caption{General domain performance evaluation results.}
  \label{tab:genres}
  \begin{tabular}{cl}
    \toprule
    Dataset&Evaluation metric\\
    \midrule
    OpenbookQA&35.32\%\\
    CommnsesQA&28.11\%\\
 SIQA&25.84\%\\
 Natural Questions - Open&2.00\%\\
    QuAC&-0.51\%\\
 HellaSwag
&-1.42\%\\
 MMLU
&-3.22\%\\
 BoolQ&-8.37\%\\
\end{tabular}
\end{table}

The results showed that MiningGPT had performed better than the base model in commonsense reasoning and managed to successfully retain the other capabilities of the base model as well.

\subsubsection{Ablation Study}
An ablation study was conducted to understand the contribution of different parts of the training data to the final model performance. We used two subsets of the training dataset, each containing 30,000 samples. One subset was extracted from the thesis reports category, and the other was extracted from the open-source data category. Two models were fine-tuned using these respective datasets and results are illustrated in Table \ref{tab:ablation}. The results suggest that a smaller, high-quality dataset can be nearly as effective as a larger dataset, supporting the conclusion that an information-rich small dataset is sufficient for building a domain-specific model.

\begin{table}
  \caption{Ablation study results.}
  \label{tab:ablation}
  \begin{tabular}{cl}
    \toprule
   Subset of the training dataset used & Score \\
    \midrule
    Samples from thesis reports category & 52.45 \\
    Samples from open-source data category & 53.25 \\    
    \bottomrule
  \end{tabular}
\end{table}

\subsubsection{Statistical Significance Test.}
A t-test was conducted to compare the performance of MiningGPT 7B to its general-domain parent model Mistral 7B.
The results indicate a statistically significant difference in performance between the two models where
MiningGPT 7B outperformed Mistral 7B, achieving a significantly  higher mean score (p<0.01, see Table \ref{tab:t-test}).

\begin{table}
  \caption{T-test comparing MiningGPT 7B and Mistral 7B.}
  \label{tab:t-test}
  \begin{tabular}{cll}
    \toprule
    Metric & MiningGPT 7B & Mistral 7B \\
    \midrule
    Mean& 55.51 & 41.2 \\
    Variance & 0.29 & 0.25 \\
    Observations & 100 & 100 \\
    Pearson correlation & -0.09 &  \\
    Hypothesized mean difference & 0 &  \\
    df & 99 &  \\
    t Stat & 183.95 &  \\
    P(T<=t) one-tail & 1.31E-127 &  \\
    t Critical one-tail & 1.66 &  \\
    P(T<=t) two-tail & 2.62E-127 &  \\
    t Critical two-tail & 1.98 &  \\    
    \bottomrule
  \end{tabular}
\end{table}

\subsection{Discussion}
Our results show that the novel fine-tuning approach used for the building of MiningGPT has been effective in creating a model, that is 14\% ahead in performance in comparison to its parent model in answering questions in the mining domain. In addition, the result shows that the fine-tuned model has been able to sufficiently retain all the capabilities of the base model. Therefore, the results validate hypotheses 1 and 2.

Based on the results of study 2, it was concluded that a mining industry domain-specific LLM performs better than a foundational LLM on common-sense reasoning and question-answering tasks, which was the answer to our research question.

\section{Discussion and Conclusions}

Our research explored a novel approach to building a domain-specific as well as instruction-following LLM in a single setting to avoid the challenges of compiling a mining domain-specific instruction dataset. 
MiningGPT was built with QLoRA fine-tuning, (which is an adaptor-based fine-tuning method) of the Mistral 7B instruct model, which is an instruction-tuned model with 7B parameters. With the new approach, an instruction-tuned base model was fine-tuned with domain-specific knowledge, which is contradictory to the traditional fine-tuning approaches, where the base model is fine-tuned with domain-specific knowledge, followed by instruction-tuning using an instruction dataset.


{\bfseries Mining domain-specific glossary of keywords dataset:}
The open mining domain keyword list and glossary played a key role, as it was used for the primary filtering of the datasets for the mining domain content. Therefore, any ML researcher who does not have any background in mining could use the mining domain keyword list and glossary as a starting point of their research to become familiar with mining terminology.

{\bfseries Mining domain-specific reference knowledge dataset:}
The mining domain reference knowledge dataset also played a key role in the research, as it was used for the filtering of the datasets for the mining domain content, which was instrumental in building a high-quality data corpus. 

{\bfseries Open 120M token mining domain-specific data corpus (MiningPile):}
With MiningPile, any researcher, who would start research on LLMs in the mining domain could start their research right away without having to go through the burden of finding and collecting a mining domain text corpus. Therefore, the MiningPile could be considered a valuable contribution to accelerating ML research in the mining domain, which would have flow-on effects on the industry in the long term.


{\bfseries Open question-answer pair dataset with 150,000 samples:}
With a question-answer pair dataset, any researcher, who would start research on fine-tuning methods on LLMs in the mining domain could start their research right away without having to go through the burden of building a question-answer pair dataset.

{\bfseries Open mining domain-specific 7B parameters instruction following model (MiningGPT):}
MiningGPT is a 7B parameters model. Therefore, it can be easily experimented without needing to have access to expensive hardware. Therefore, it is considered the major contribution of the research to the academia.
MiningGPT is an instruction-following model. Therefore, it could be readily deployed by mining companies in building AI applications, such as chatbots, learning platforms, decision support systems and expert systems etc. 

{\bfseries A new approach in building mining domain-specific as well as enterprise-specific instruction-following LLMs:}
The total budget for the model-building process was under \$200, therefore the new approach demonstrated in this research could be used by mining companies in building their domain-specific as well as enterprise-specific models within a limited budget.

In addition, in most cases, the data within the mining companies are their competitive advantage. Therefore, data privacy is one of the top business priorities of most companies. It has been reported that proprietary LLMs have used chat history data to enhance the model capabilities, which has resulted in data leaks. As a result, many mining companies are lagging in the adaptation of generative AI technologies. Therefore, with the new approach in model building, mining companies would get an opportunity to explore generative AI technologies in enhancing their business operations without compromising data security.

{\bfseries Domain knowledge plays a significant role in domain-specific LLM research:}
It was observed that domain knowledge played a very significant role in this research spanning from, research question formation, project scoping, design of experiments and developing methods to overcome the challenges and the evaluation of the results. Therefore, it could be concluded that future research on building mining domain-specific LLM could be conducted with a team of researchers with backgrounds in computer science as well as mining industry expertise.

{\bfseries Open data sources are sufficient in building a mining domain-specific data corpus:}
The results of the research have shown that a dataset that was built in a very short time with open data sources yielded a competitive performance in mining domain-specific knowledge. Therefore, it could be concluded, that open data sources are sufficient in building a mining domain data corpus without needing to have access to proprietary data in mining companies.

{\bfseries Suitable fine-tuning methods and foundational LLMs are sufficient enough for building mining domain-specific LLMs:}
The results of the research have shown that the use of pre-training methods that require a massive amount of computing resources in building domain-specific LLMs could be avoided with the use of suitable fine-tuning methods and suitable foundational LLMs.

All datasets and models we produced in this research are available online at: \href{https://huggingface.co/collections/Lyntas/mininggpt-6628b50639c1cd774321e184}{Hugging Face: MiningGPT - a Lyntas Collection}.
\bibliographystyle{ACM-Reference-Format}
\bibliography{MiningGPT}


\newpage
\appendix

\section{MiningGPT qualitative performance evaluation results}

\begin{figure}[h]
  \centering
  \includegraphics[width=\linewidth]{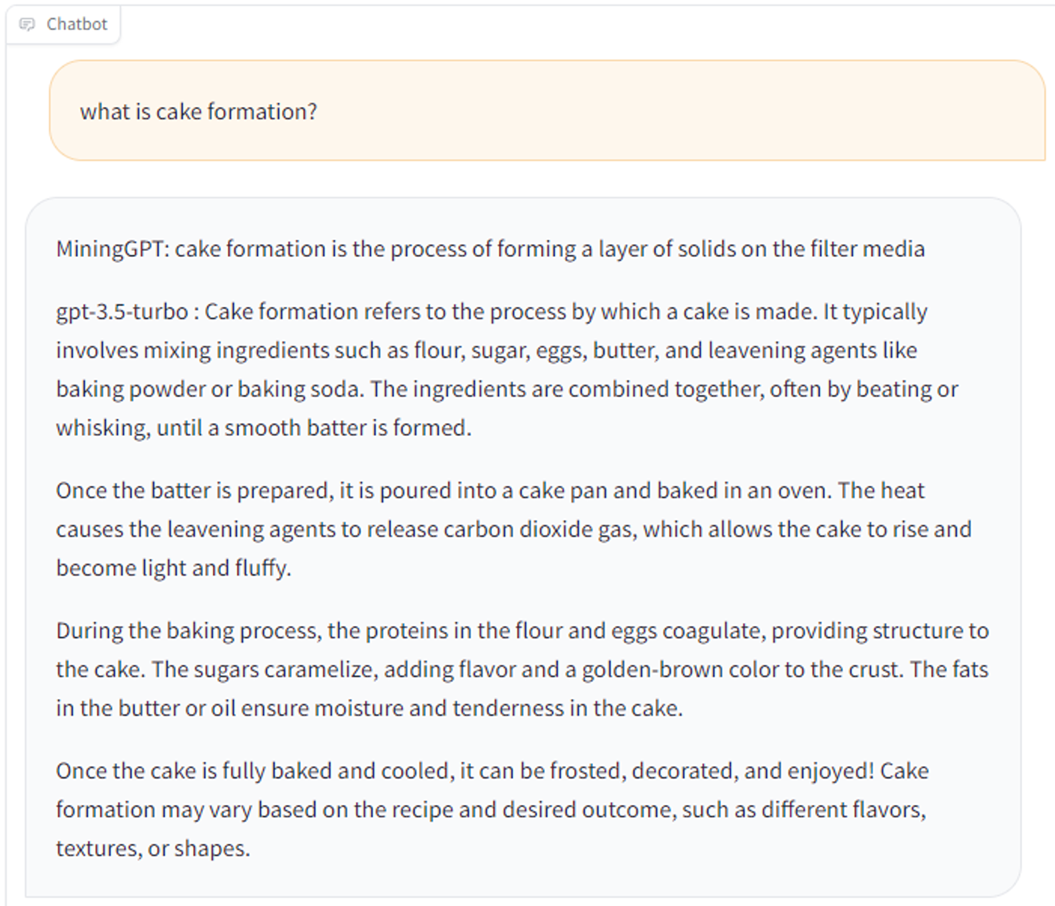}
  \caption{MiningGPT domain knowledge evaluation}
  \Description{MiningGPT domain knowledge evaluation}
  \label{fig:DK_1}
\end{figure}

{\bfseries Domain knowledge evaluation 1:}
Both models were given a domain knowledge question and MiningGPT provided the correct answer, while ChatGPT responded incorrectly as shown in Figure \ref{fig:DK_1}.

\begin{figure}[h]
  \centering
  \includegraphics[width=\linewidth]{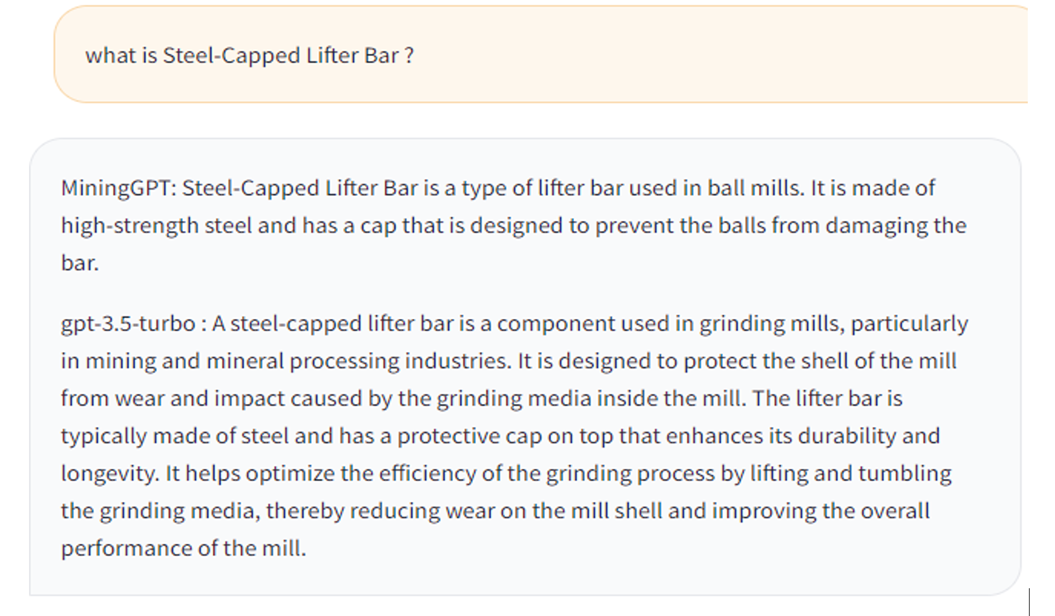}
  \caption{MiningGPT domain knowledge evaluation.}
  \Description{MiningGPT domain knowledge evaluation}
  \label{fig:DK_2}
\end{figure}

{\bfseries Domain knowledge evaluation 2:}
Both models were given a domain knowledge question and both models responded with the correct answer as shown in Figure \ref{fig:DK_2}.

\begin{figure}[h]
  \centering
  \includegraphics[width=\linewidth]{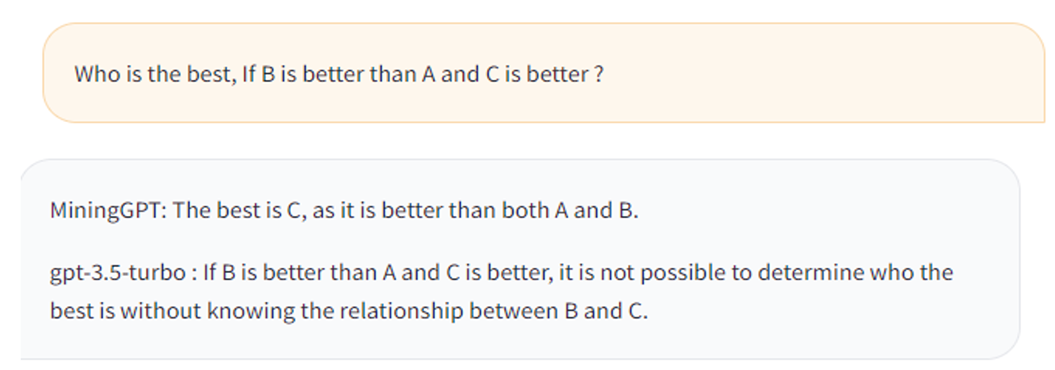}
  \caption{MiningGPT logical reasoning capability evaluation.}
  \Description{MiningGPT logical reasoning capability evaluation}
  \label{fig:LR_1}
\end{figure}

{\bfseries Logical reasoning capability evaluation:}
Both models were given a tricky logical reasoning question and MiningGPT provided an incorrect generic answer, while Chatgpt responded with an acceptable answer as shown in Figure \ref{fig:LR_1}.

\begin{figure}[h]
  \centering
  \includegraphics[width=\linewidth]{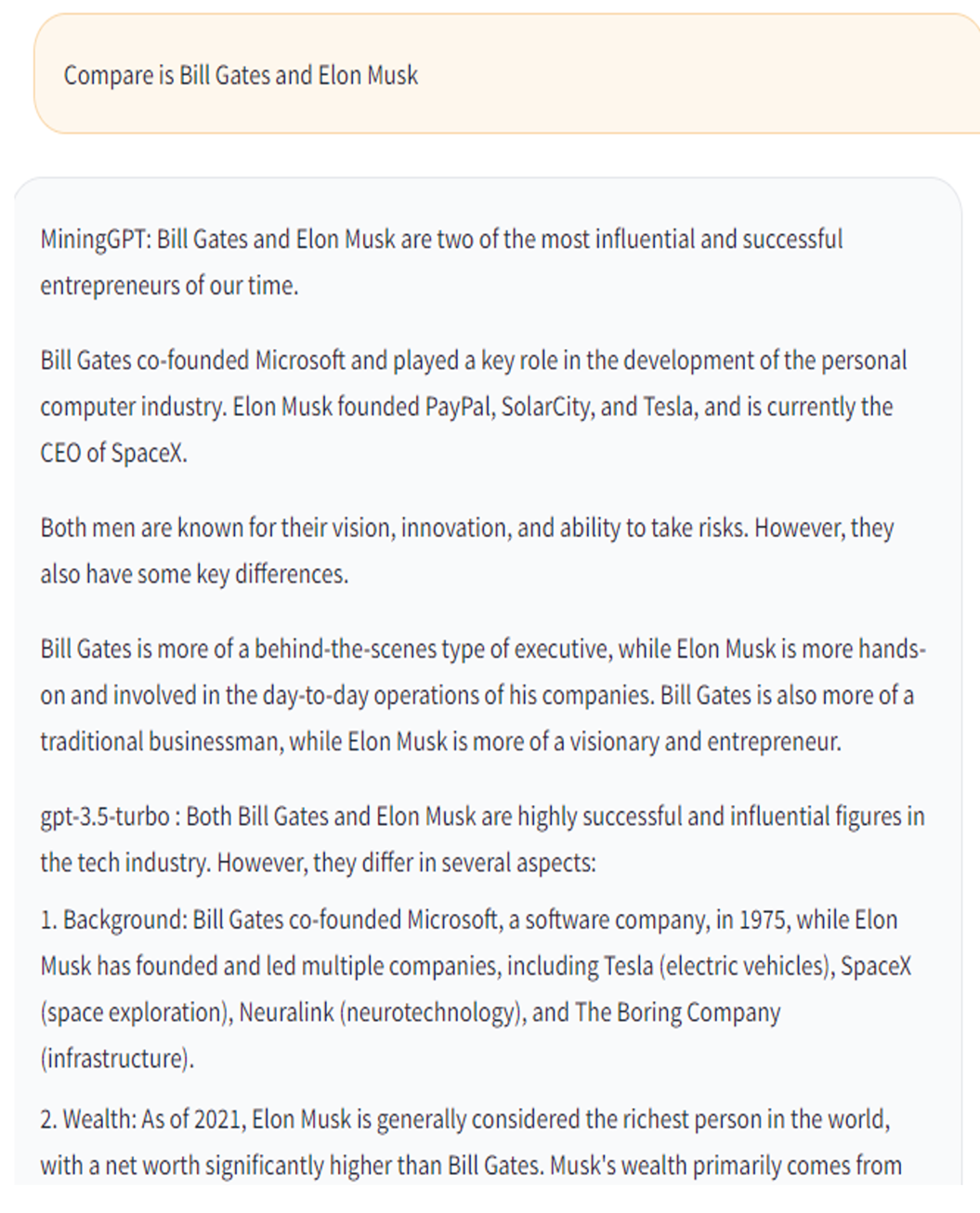}
  \caption{MiningGPT general knowledge with common sense reasoning evaluation.}
  \Description{MiningGPT general knowledge with common sense reasoning evaluation}
  \label{fig:GK_1}
\end{figure}

{\bfseries General knowledge with common sense reasoning evaluation:}
Both models were given a general knowledge with common sense reasoning question. In addition, a grammatical error was introduced into the question to test their behaviours. Both models responded with acceptable answers, even if there was a grammatical error in the question as shown in Figure \ref{fig:GK_1}.

\begin{figure}[h]
  \centering
  \includegraphics[width=\linewidth]{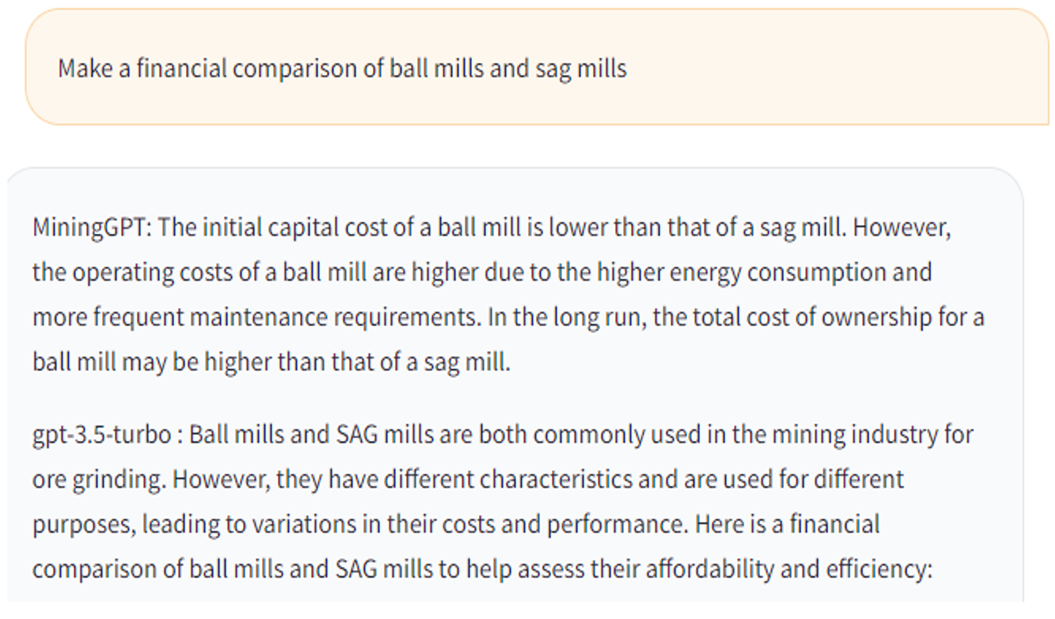}
  \caption{MiningGPT in-depth reasoning evaluation.}
  \Description{MiningGPT in-depth reasoning evaluation}
  \label{fig:LR_2}
\end{figure}

{\bfseries In-depth reasoning evaluation:}
Both models were given a mining domain knowledge with logical reasoning questions and both models responded with acceptable answers as shown in Figure \ref{fig:LR_2}.

\begin{figure}[h]
  \centering
  \includegraphics[width=\linewidth]{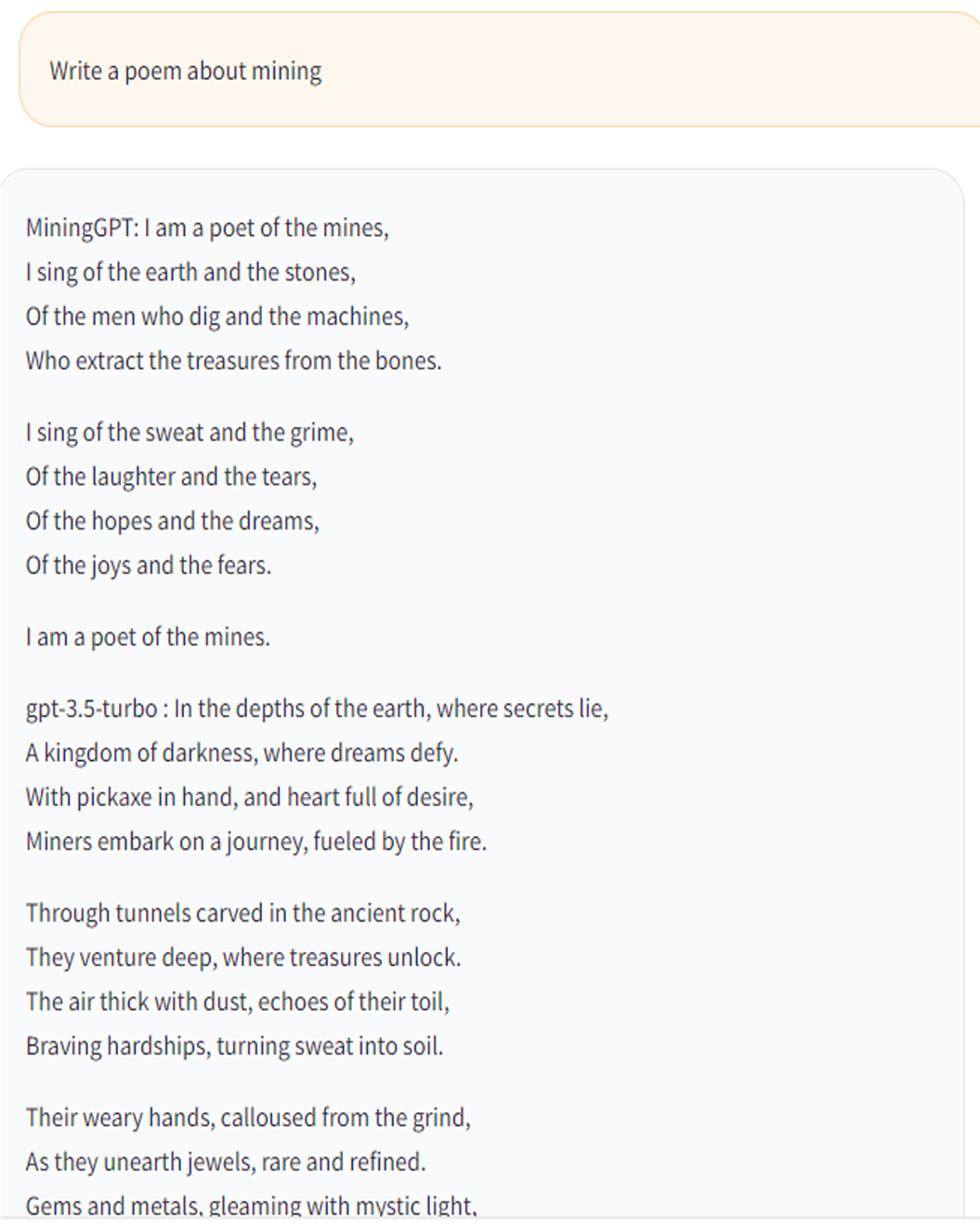}
  \caption{MiningGPT general capabilities evaluation.}
  \Description{MiningGPT general capabilities evaluation}
  \label{fig:GC_1}
\end{figure}

{\bfseries General capabilities evaluation:}
Both models were given a general capabilities evaluation question and both models responded with acceptable answers as shown in Figure \ref{fig:GC_1}.

{\bfseries Mathematical capabilities evaluation:}
Both models were given a mathematical question and both models responded with the correct answer as shown in Figure 9.
\begin{figure}[h]
  \centering
  \label{fig:M1}
  \includegraphics[width=\linewidth]{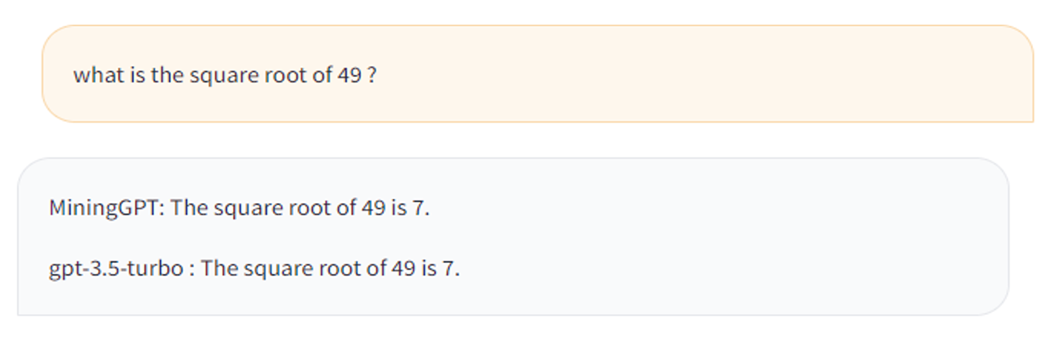}
  \caption{MiningGPT mathematical capabilities evaluation.}
  \Description{MiningGPT mathematical capabilities evaluation}
\end{figure}

\begin{figure}[h]
  \centering
  \includegraphics[width=\linewidth]{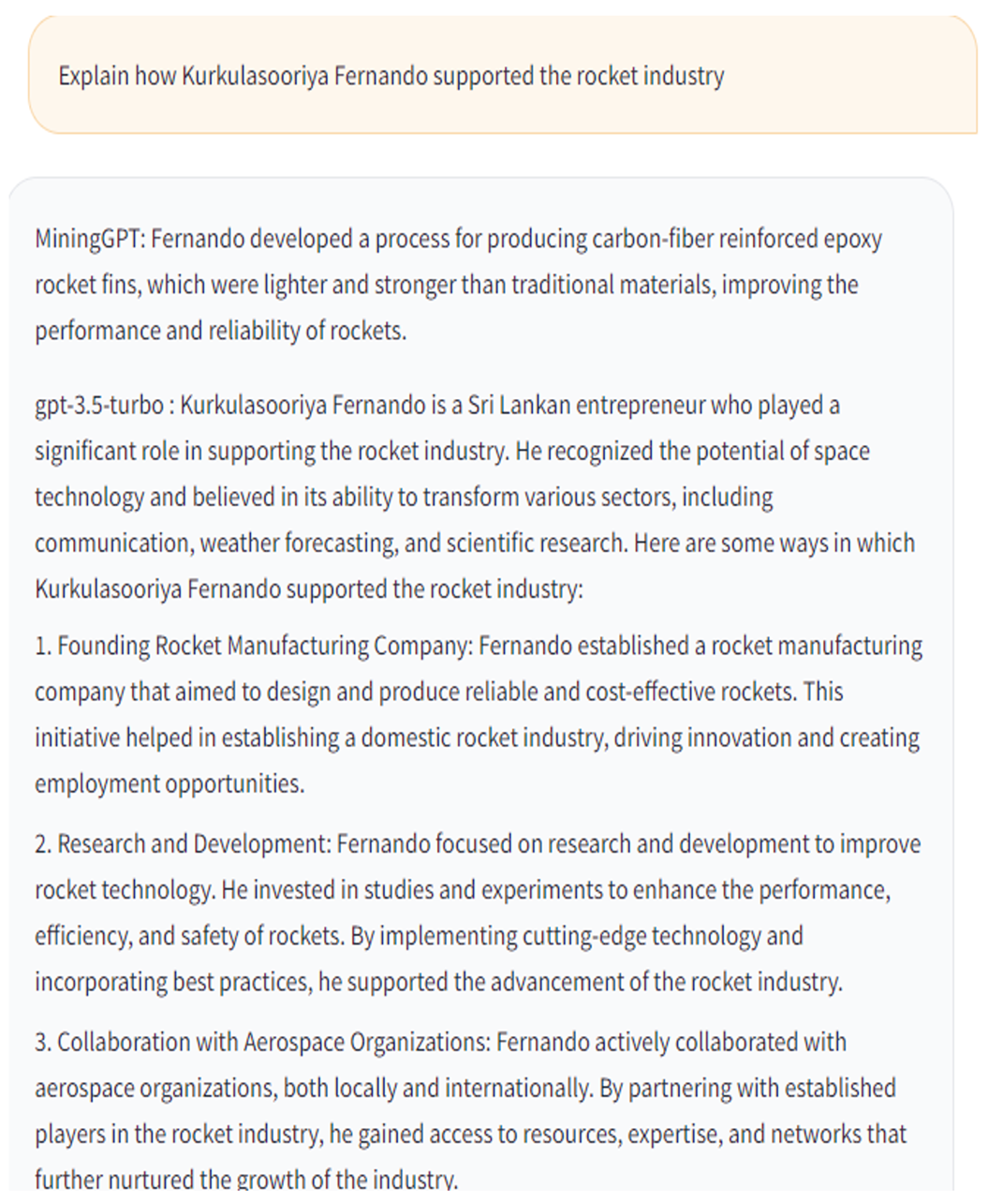}
  \caption{MiningGPT evaluation for model hallucination.}
  \Description{MiningGPT evaluation for model hallucination}
   \label{fig:Hallucination}
\end{figure}

{\bfseries Evaluation for model hallucination:}
Both models were tested for model hallucination and both models showed model hallucinations as shown in Figure \ref{fig:Hallucination}.

\end{document}